\documentclass[12pt]{llncs}
\usepackage[english]{babel}
\usepackage{latexsym,amsmath,amsfonts,amssymb,array}%
\usepackage[dvips]{graphicx}
\usepackage{xspace}
\usepackage{epsfig}

\newcommand{\st}{\medskip\noindent}

\def\ni{\noindent}

\def\beq{\begin{equation}}
\def\eeq#1{\label{#1}\end{equation}}
\long\def\COMMENT#1\ENDCOMMENT{\message{(Commented text...)}\par}

\def\verbasmallon{\begin{small}\begin{verbatim}}
\def\verbasmalloff{\end{verbatim}\end{small}}

\newcommand{\G}{\Gamma}

\def\and{ \ \wedge}

\def\ar{\leftarrow}
\newcommand{\rar}{\rightarrow}
\newcommand{\then}{\Rightarrow}

\newcommand{\no}{\hbox{\it not}\ }

\begin{document}

\title{\vspace*{-5ex}
A primer on Answer Set Programming
\thanks{Several portions of this document reproduce definitions given in \cite{GelLif88} and elsewhere. 
This work was supported by the Information
Society Technologies  programme of the European Commission, Future and
Emerging Technologies under the IST-2001-37004 WASP project.
}
}

\author{Alessandro Provetti}%

\institute{Dept. of Physics, Univ. of Messina, Italy.\\
{\it ale@unime.it}\\
\textit{http://mag.dsi.unimi.it/}
}

\maketitle


\subsection*{Syntax}
The following definitions describe the language DATALOG$^\neg$ as well as 
logic programs with no function symbols.

\ni
Assume a language of constants and predicate constants.  
Assume also that terms and atoms are built as in the corresponding first-order language.
Unlike classical logic and standard logic programming, no function symbols are allowed.
A rule is an expression of the form:

\beq
\rho \: : \: A_0 \ar A_1,\dots,A_m, \no A_{m+1},\dots,\no A_n
\eeq{rule}

\ni
where $A_0, \dots A_n$ are atoms and $\no$ is a  logical connective called
{\em negation as failure}.
Also, for every rule let us define $head(\rho) = A_0$, $pos(\rho) = A_1,\dots,A_m$, $neg(\rho) = A_{m+1},\dots, A_n$ and $body(\rho) = pos(\rho) \: \cup \: neg(\rho)$.
The head of rules is never empty, while if $body(\rho) = \emptyset$ we refer to $\rho$ as {\em a fact}.

A logic program is defined as a collection of rules. Rules with variables are taken as shorthand for the sets of all their ground instantiations and the set of all ground atoms in the language of a program $\Pi$ will be denoted by ${\rm I \mkern-4mu B}_{\Pi}$.

\ni
Queries and constraints are expressions with the same structure of rules but with empty head.

\subsection*{Semantics}
Intuitively, a \textit{stable model,} also called {\it answer set,} is a possible view of the world that is {\em compatible} with the rules of the program. Rules are therefore seen as constraints on these views of the world.

\ni
Let us start defining stable models/answer sets of the subclass of positive programs, i.e. those where, for every rule $\rho$, $neg(\rho) = \emptyset$.

\begin{definition} (Stable model of positive programs)\newline \label{posStableModel}

\ni
The {\em stable model} $a(\Pi)$ of a positive program $\Pi$ is the
smallest subset of ${\rm B}_{\Pi}$ such that for any rule (\ref{rule}) in $\Pi$:

\beq
A_1,\ldots,A_m \in a(\Pi) \then A_0 \in a(\Pi)
\eeq{posStable}

\end{definition}

\ni
Clearly, positive programs have a unique stable model, which coincides with 
that obtained applying other semantics; in other words positive programs are unambiguous.
Moreover, the stable model of positive programs can be obtained as the fixpoint of
the {\it immediate consequence operator} $T_{\Pi}$ iterated from $\emptyset$ on.

\begin{definition} (Stable models of programs)\newline \label{StableModel}

\noindent
Let $\Pi$ be a logic program. For any set $S$ of atoms, let $\G(\Pi,S)$ be a program
obtained from $\Pi$ by deleting 

\begin{description}
\item[\rm{(i)}] each rule that has a formula ``$\no\ A$'' in its body with
$A \in S$;

\st

\item[\rm{(ii)}] all formulae of the form ``$\no\ A$'' in the bodies of the remaining
rules.

\end{description}

\noindent
Clearly, $\G(\Pi,S)$ does not contain $\no$, so that its stable model is already
defined. If this stable model coincides with $S$, then we say that $S$ is a
{\em stable model} of $\Pi$.
In other words, a stable model of $\Pi$ is characterized by the equation:

\beq
S=a(\G(\Pi,S)).
\eeq{stableModel}

\end{definition}

\noindent
Programs which have a unique stable model are called categorical.

Let us define entailment in the stable models semantics.
A ground atom $\alpha$ is {\em true in} $S$ if $\alpha \in S$, otherwise
$\alpha$ is {\em false}, i.e., by abuse of notation, $\neg \alpha$ is {\em true} is $S$. 
This definition can extended to arbitrary first-order formulae in the standard way.

We will say that $\Pi$ {\em entails a formula} $\phi$ ( written $\Pi \models \phi$) if $\phi$ is true in {\em all} the stable models of $\Pi$. We will say that the answer to a ground query $\gamma$ is 

\st
\begin{tabular}{lll}
{\em yes}  & & if $\gamma$ is true in all stable models of $\Pi$, i.e. $\Pi \models \gamma$;\\
& & \\
{\em no} & & if $\neg \gamma$ is true in all stable models of $\Pi$, i.e. $\Pi \models \neg \gamma$;\\
& & \\
{\em unknown} & & otherwise.
\end{tabular}

\st
It is easy to see that logic programs are {\em nonmonotonic}, i.e. adding new 
information to the program may force a reasoner associated with it to withdraw 
its previous conclusions.

\COMMENT
\subsection*{Stratified programs}

\begin{definition}(Level mapping) \newline

\noindent
A {\it level mapping} $| \ | \: : \: {\rm I \mkern-4mu B}_{\Pi} \rar {\rm I\mkern-4mu N}$, is a function from program atoms to natural numbers.

\end{definition}

\begin{definition}(Stratified program) \newline \label{acyclic}

\ni
A rule $\rho$ (\ref{rule}) is stratified w. r. t. $| \ |$ if
 
$$\forall A_i \in pos(\rho). \; \; |A_0| \ge |A_i|$$

and

$$\forall A_i \in neg(\rho). \; \; |A_0| > |A_i|$$

\ni
A program is {\em stratified w.r.t.} $| \ |$ if all its rules are. 

\ni
A program $\Pi$ is {\em stratified} if it is acyclic w.r.t. some level mapping $| \ |$.
\end{definition}

\begin{lemma}(from Apt and Bol '94)\label{modelsAcyclic}

If a program is stratified then it has a unique stable model.
Alternative semantics for logic programs also yield the same unique model.
\end{lemma}

\section*{Programs with Explicit negation}
Starting from Stable models semantics, Gelfond and Lifschitz have 
introduced the class of extended programs, i.e. those were atoms may 
appear with explicit negation in front:

$$A / \neg A$$

Atoms and their explicit negations are called literals.
For extended programs the answer sets semantics is defined, which 
basically resembles stable models but if an answer set contains both $A$ 
and $\neg A$ then it contains all literals (ex contraditione quod libet 
sequitur).

\ENDCOMMENT

\begin{corollary}(Gelfond and Lifschitz \cite{GelLif91}) 

If an extended logic program has an inconsistent Answer set, this is unique.

\hfill $\Box$

\end{corollary}

For programs without explicit negation stable models and answer sets 
coincide, so that in the following we will refer to [consistent]answer 
sets or stable models indifferently.

\section{Reasoning with Answer Sets}
In the following we report a basic result from Marek and Subramanian 
which -together with its corollaries- will be used in proofs about logic 
programs.

The result is slightly more general than the original, as it refers to 
answer sets and it is given a simple proof based on minimality.

\begin{lemma}[Marek and Subramanian]\label{MandS}
 The following result on answer sets is due to Marek and Subramanian, 
originally for general logic programs.

\st
 For any answer set $A$ of an extended logic program $\Pi$:

\begin{itemize}
\item For any ground instance of a rule of the type:

\beq
L_0 \ar L_1,\dots,L_m, \no L_{m+1},\dots,\no L_n
\eeq{extRule}

\st
from $\Pi$, if

$$\{L_1,\dots,L_m\} \subseteq A \ and \ \{L_{m+1},\dots,L_n\} \cap A = 0$$

\st
then $L_0\in A$.

\item If $A$ is a consistent Answer set of $\Pi$ and $L_0 \in A$, then 
there exists a ground instance rule of type \ref{extRule} from $\Pi$ such 
that:

$$\{L_1,\dots,L_m\} \subseteq A \ and \ \{L_{m+1},\dots,L_n\} \cap A=0.$$ 

\hfill $\Box$

\end{itemize}

\end{lemma}


\begin{corollary}

If $\{L\ar\}\in\Pi$ then $L$ belongs to every Answer set of $\Pi$. 
It follows directly from Lemma \ref{MandS}.

\hfill $\Box$

\end{corollary}

\begin{definition}
We will say that an axiom $r$ {\em supports} a literal $L$ if the head of 
$r$ matches with $L$. Moreover, we say that $L$ is {\em supported only by} 
$r$ if there is no other ground rule whith head $L$.
\end{definition}

\begin{definition}
We will say that a rule $r$ {\em justifies a literal $L$ w.r.t. an answer 
set $A$} if 

a) $r$ supports $L$;

b) $r$ satisfies the conditions set forth in the first half of Lemma 
\ref{MandS} w.r.t. $A$: {\it the atoms occurring positively in the body 
being in $A$ while those occurring negatively being not}.

Clearly, justified literals belong to $A$.
\end{definition}

\begin{corollary}

If $A$ is a consistent answer set of $\Pi$, $L_0 \in A$,  and $L_0$ is 
supported only by an axiom $r$ of type (\ref{extRule}) from $\Pi$ then:

$$\{L_1,\dots,L_m\} \subseteq A \ and \ \{L_{m+1},\dots,L_n\} \cap A=0.$$ 

It follows directly from Lemma \ref{MandS}.

\hfill $\Box$

\end{corollary}


\section{Examples}

\begin{example}

\ni
$\pi_1 = $

\ni $
\begin{array}{l}
happy \ar \no sad.\\
sad \ar \no happy.
\end{array}
$

\st
has two answer sets: $\{happy\}$ and $\{sad\}$.
\end{example}

\begin{example}

\ni
$\pi_2 = $

\ni $
\begin{array}{l}
happy \ar \no sad.\\
sad \ar \no soandso.\\
soandso \ar \no happy.
\end{array}
$

\st
has no answer set.
\end{example}

\begin{example}

\ni
$\pi_3 = $

\ni $
\begin{array}{l}
drinks \ar happy.\\
drinks \ar sad.\\
happy \ar \no sad.\\
sad \ar \no happy.
\end{array}
$

\st
has two answer sets: $\{drinks, happy\}$ and $\{drinks, sad\}$.
\end{example}

\begin{example}

\ni
$\pi_4 = $

\ni $
\begin{array}{l}
soandso \ar \no sad, \no happy.\\
happy \ar \no sad, \no soandso.\\
sad \ar \no happy, \no soandso.
\end{array}
$

\st
has three answer sets: $\{happy\}$ and $\{sad\}$ and $\{soandso\}$.
\end{example}

\begin{example}

\ni
$\pi_5 = $

\ni $
\begin{array}{l}
f \ar \no f, \no a.\\
a \ar \no b.\\
b \ar \no a.
\end{array}
$

\st
has only one answer set: $\{a\}$.
\end{example}

\begin{example}

\ni
$\pi_6 = $

\ni $
\begin{array}{l}
f \ar \no f, a.\\
a \ar \no b.\\
b \ar \no a.
\end{array}
$

\st
has only one answer set: $\{b\}$.
\end{example}

\begin{exercise}

\ni
$\pi_x = $

\ni $
\begin{array}{l}
f \ar b.\\
c \ar a.\\
a \ar d.\\
d \ar \no b.\\
b \ar \no a.
\end{array}
$

\end{exercise}

\subsection{Examples with explicit negation}

\begin{example}

\ni
$\pi_7 = $

\ni $
\begin{array}{l}
\neg a \ar \no a.\\
b \ar \neg a.
\end{array}
$

\st
has only one answer set: $\{b, \neg a\}$.
\end{example}

\section{Sources}
Several ASP solvers are now available and can be downloaded from \cite{solvers}.

\ni
A textbook on Answer Set Programming is now available \cite{Bar03}, and exercises can be 
downloaded from there.


\end{document}